\documentclass[letters]{fcs}
\usepackage{layout}
\usepackage{subfig}
\usepackage{hyperref}
\usepackage{graphicx} 

\usepackage{soul}
\usepackage{xcolor}

\title{ECG-Mamba-V2: Architectural Refinements to a Bidirectional State Space Model for Multi-Label 12-Lead ECG Classification}
\author[1]{Huawei~JIANG}
\author[2]{Husna~MUTAHIRA}
\author[3]{Shibo~WEI}
\author[4]{Gan~HUANG}
\author[5]{Vladimir~SHIN}
\author[3]{Dongryeol~RYU}
\author[6,+]{Juneho~YI}
\author[7,+]{Mannan Saeed~MUHAMMAD}
\address[1]{Department of Computer Science and Engineering, Sungkyunkwan University, Suwon, 16419, South Korea}
\address[2]{Department of Computer Science and Engineering, Sogang University, Seoul, 04104, South Korea}
\address[3]{Department of Biomedical Science and Engineering, Gwangju Institute of Science and Technology, Gwangju, 61005, South Korea}
\address[4]{Department of Computer Science, Nusa Putra University, Sukabumi, 43152, Indonesia}
\address[5]{Department of Artificial Intelligence, Financial University under the Government of the Russian Federation, Moscow, 125167, Russia}
\address[6]{Department of Electrical and Computer Engineering, Sungkyunkwan University, Suwon, 16419, South Korea}
\address[7]{School of Electronic Engineering and Computer Science, Queen Mary University of London, London, E1 4NS, United Kingdom}

\corremail{jhyi@skku.edu; mannan.muhammad@qmul.ac.uk}
\fcssetup{
  received = {month dd, yyyy},
  accepted = {month dd, yyyy},
}

\makeatletter
\expandafter\def\csname g__fcs_abstract_tl\endcsname{}
\expandafter\def\csname l__fcs_keywords_tl\endcsname{}
\makeatother

\begin{document}

\section{Introduction}
Cardiovascular diseases remain a leading cause of global mortality, necessitating efficient and accurate tools for electrocardiogram (ECG) analysis \cite{11176036}. While deep learning architectures like Transformers have improved automated detection, they often struggle with the high computational cost and long-range dependencies inherent in multi-lead ECG recordings \cite{natarajan2020wide, nejedly2021classification}. 

\begin{figure}
    \centerline{\includegraphics[width=\columnwidth]{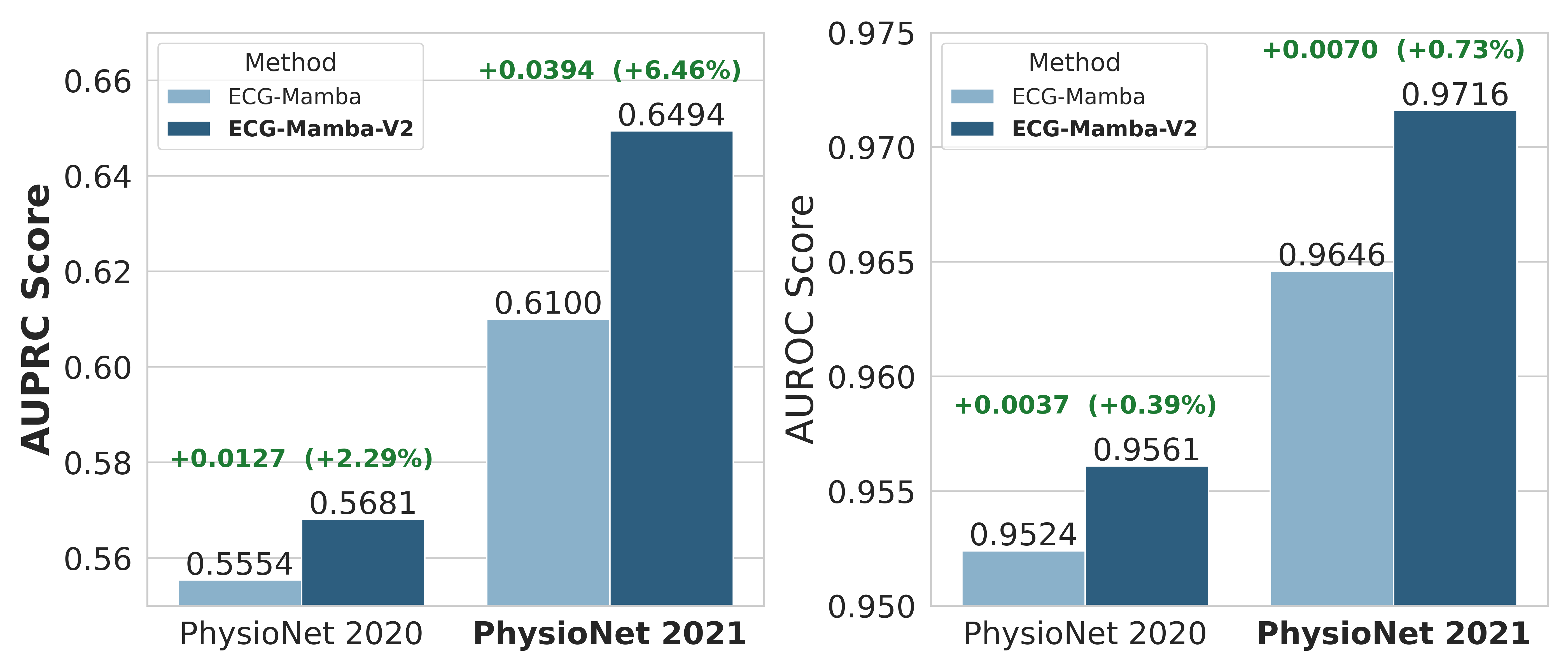}}
    \caption{Performance comparison of ECG-Mamba and ECG-Mamba-V2 on the PhysioNet/CinC Challenge 2020 and 2021 datasets. As the 2021 dataset is a superset of the 2020 dataset and roughly twice its size, its results are the more reliable indicator. 
    }
    \label{sec:comparison}
\end{figure}

State Space Models (SSMs), particularly Mamba \cite{gu2024mamba}, address these challenges by providing linear scaling and global receptive fields for sequential data. 
ECG-Mamba-V2 builds on Vision Mamba (Vim) \cite{zhu2024vision} and refines the original ECG-Mamba \cite{11176036}.
Figure \ref{sec:comparison} illustrates the substantial Macro AUPRC and Macro AUROC increases achieved by ECG-Mamba-V2. The updated architecture extracts temporal features from standardized ECG signals using a one-dimensional convolutional neural network (CNN) layer. Crucially, several empirical modifications distinguish the new Vim encoder from its predecessor. 
The class token is appended at the end of the token sequence rather than inserted at the midpoint. Dropout is applied at a uniform rate of 0.1 in every block. In the bidirectional mixer, the forward and backward scan outputs are summed rather than averaged before the output projection.


Experimental results on the PhysioNet/CinC 2020 and 2021 benchmark datasets show that ECG-Mamba-V2 achieves consistently better performance than existing methods \cite{natarajan2020wide, nejedly2021classification, elyamani2024deep}, including the official challenge winners and the earlier ECG-Mamba model \cite{11176036}. 
The improvement holds across both benchmarks and across all 15 training runs per configuration. Code is available at \cite{source_code}.

The key contributions of this work include:
\begin{itemize}
    \item Class token position matters more than encoder capacity in Vim-based encoders. Prior models \cite{zhu2024vision, erol2024audio, 11176036} place the token at the midpoint; this work places it at the end.
    \item Refinements to class token placement, mixer scaling, and dropout raise Macro-AUPRC from 0.6100 to 0.6494 over the predecessor, at 34\% fewer parameters and 38\% higher throughput.
    \item A proxy patient key groups related recordings into one fold, addressing a leakage risk left open by prior work \cite{leinonen2024empirical}.
\end{itemize}

\begin{figure}[!t]
    \centerline{\includegraphics[width=0.5\textwidth]{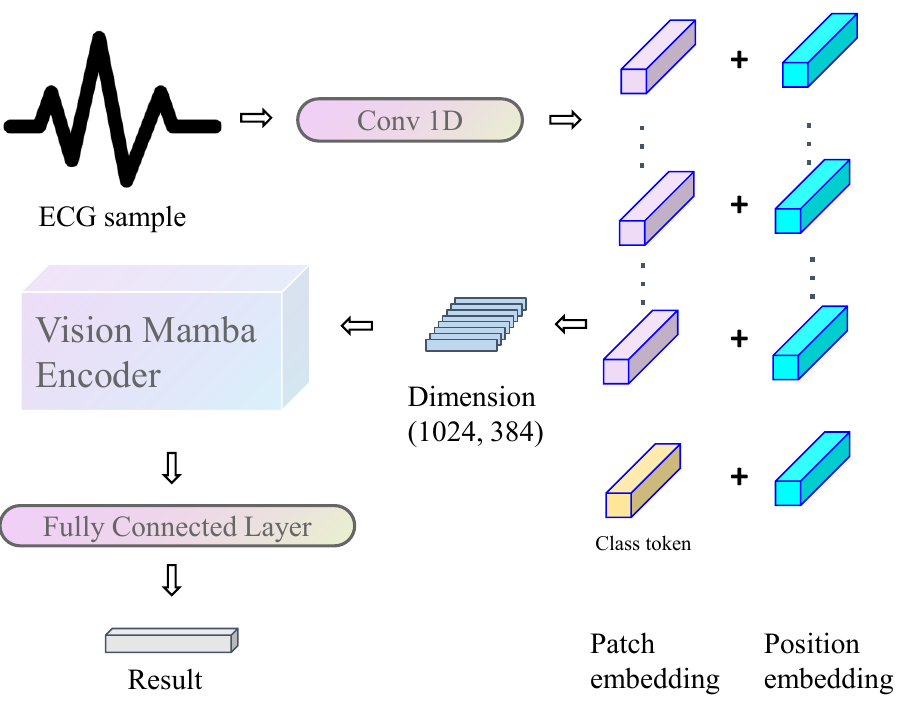}}
    \caption{The ECG-Mamba-V2 architecture.}
    \label{Figure-ECG-Mamba}
\end{figure}

\section{Algorithm architecture}

\subsection{ECG Sample Pre-Processing}
All ECGs are standardized to 500 Hz using polyphase downsampling for 1000 Hz data and Fast Fourier Transform upsampling for 257 Hz data. The framework then follows \cite{nejedly2021classification, 11176036} to fix the signal length at 8192 samples by applying right zero-padding or random cropping.

\subsection{ECG-Mamba-V2}
\label{sec_ECG-Mamba-V2}
Vim was developed for two-dimensional visual data; this work adapts it to 12-lead ECG signals, each represented as a $12 \times N$ tensor for 12 leads and $N$ time samples. A one-dimensional CNN layer placed before the Vim encoder projects the 12-channel input into 384 channels using a kernel size of 16 and a stride of 8.
This process extracts temporal features in a form compatible with the Vim encoder.
The resulting feature map from the CNN projection is partitioned into patches of shape $(1023, 384)$. A class token is appended to the end of the token sequence. This differs from Vim \cite{zhu2024vision}, Audio Mamba \cite{erol2024audio}, and ECG-Mamba \cite{11176036}, all of which insert the token at the sequence midpoint. This addition yields a final sequence length of $(1024, 384)$. Positional embeddings are then added to the patch sequence based on the approach in \cite{zhu2024vision}.

The Vim encoder incorporates several empirical modifications. First, dropout is applied at a uniform rate of 0.1 in every block rather than at a rate increasing linearly with depth, which would leave the earliest blocks almost unregularized; the uniform setting distributes regularization evenly across the encoder. Second, the forward and backward scan outputs are summed and passed directly to the output projection, without the 1/2 scaling of the original formulation. Since the block output is added to the residual path, this factor halves the mixer's contribution relative to the residual, and removing it lets both directional representations contribute at full magnitude. 
Third, the fused residual addition and normalization operation is disabled (\texttt{fused\_add\_norm=False}) for reproducibility across hardware; the fused and sequential paths compute the same function.


These adaptations collectively constitute a configuration distinct from those in \cite{zhu2024vision, 11176036} to reflect optimizations specific to ECG signal processing. 
An overview of the overall architecture is illustrated in Fig. \ref{Figure-ECG-Mamba}. The best performance was achieved by fine-tuning a model with 16 Vim encoder blocks.

\section{Experimental setup}

Experiments were conducted in PyTorch on an NVIDIA RTX 3090 Ti. The primary data source was the PhysioNet/CinC 2021 dataset \cite{PhysioNet-challenge-2021-1.0.3}, which comprises 88{,}253 recordings across 26 diagnostic categories. 
Recording JS23074 (Ningbo First Hospital) was excluded because the official release provides no header file with its labels, leaving 88{,}252 recordings for all experiments.
The PhysioNet/CinC 2020 dataset (27 categories) was evaluated as a second benchmark. 
To mitigate data leakage, data partitioning followed a group-stratified 5-fold cross-validation scheme. As the public release does not provide patient identifiers, exact patient-level grouping is not achievable; recordings sharing the same source database, age, gender, and diagnostic label were therefore treated as a proxy for a single patient and constrained to the same fold.
This yielded an 80/20 train/validation ratio per fold. 
Every configuration was trained under three random seeds ($0$, $1$, and $2$), so each reported value is the mean over 15 runs ($3$ seeds $\times$ $5$ folds). 
Training employed the Adam optimizer with a cosine annealing learning rate schedule consisting of a 5-epoch linear warm-up to a peak of $6 \times 10^{-4}$, followed by a 13-epoch decay. Following ECG-Mamba \cite{11176036}, binary cross-entropy loss was used as the training objective.  
Finally, macro-averaged AUPRC and AUROC were used to evaluate multi-label classification performance under class imbalance. 
Both metrics are threshold-free, summarizing performance across all operating points, and were computed by averaging over the diagnostic classes within each run and then over the 15 runs.

\section{Results and analysis}
This study evaluated ECG-Mamba-V2 against the leading Challenge entries: the Transformer-based model of team Prna \cite{natarajan2020wide}, first in 2020, and the ResNet-based model of team ISIBrno \cite{nejedly2021classification}, first in 2021, together with a 2D-CNN image-representation model \cite{elyamani2024deep} and the original ECG-Mamba \cite{11176036}. Hyperparameters followed the original implementations. 
All models except Prna shared an identical preprocessing pipeline; Prna retained its own\cite{natarajan2020wide}, as its wide-and-deep design is coupled to a model-specific feature extraction stage. All models were evaluated on identical fold assignments.

Table~\ref{table_ECG-mamba-V2-2021} reports accuracy and computational cost on Challenge 2021. ECG-Mamba-V2$^{ud}$, applying uniform dropout of 0.1 in every block, leads on both metrics at 0.6494 macro AUPRC and 0.9716 macro AUROC, exceeding ECG-Mamba$^{d}$ by 0.0394 and 0.0070 with 34\% fewer parameters (17.17 M against 26.09 M), 9\% fewer MACs, and 38\% higher throughput (252 against 183 f/s); cost therefore falls as accuracy rises. The architectural gain is separable from regularization, as the model without dropout already reaches 0.6333. 
A Wilcoxon signed-rank test over the 15 paired runs confirms the improvement, with wins in all 15 runs on both metrics (median 0.0405 macro AUPRC and 0.0067 macro AUROC; $W = 0$, $p = 6.1 \times 10^{-5}$).

\begin{table}[htbp]
\centering
\caption{Performance comparison on PhysioNet/CinC Challenge 2021. Abbreviations: Params (Parameters), MACs (Multiply-Accumulate Operations), T-put (Throughput in file/s, computed with a batch size of 20). Superscripts denote variants: $d$, dropout increasing linearly from 0 to 0.1 across blocks; $ud$, uniform dropout of 0.1 in every block; $t$, the tiny configuration (4 blocks, model dimension 128, input length 1000). Values are the mean over 15 runs (3 seeds $\times$ 5 folds) with the standard deviation in subscript.}
\label{table_ECG-mamba-V2-2021}
\resizebox{\columnwidth}{!}{
\begin{tabular}{llll|ll}
\toprule
\textbf{Method} & \textbf{Params} & \textbf{MACs} & \textbf{T-put (f/s)} & \textbf{AUPRC} & \textbf{AUROC} \\ 
\midrule
ISIBrno & \multicolumn{1}{r}{6.56 M}  & \multicolumn{1}{r}{5.19 G} & \multicolumn{1}{r|}{2419} & 0.5170$_{\pm 0.0014}$ & 0.9073$_{\pm 0.0019}$ \\
2D-CNN & \multicolumn{1}{r}{0.59 M}  & \multicolumn{1}{r}{7.47 G} & \multicolumn{1}{r|}{4913} & 0.5431$_{\pm 0.0030}$ & 0.9389$_{\pm 0.0008}$ \\

ECG-Mamba & \multicolumn{1}{r}{26.09 M} & \multicolumn{1}{r}{19.22 G} & \multicolumn{1}{r|}{183} & 0.6083$_{\pm 0.0002}$ & 0.9643$_{\pm 0.0001}$ \\
ECG-Mamba$^{d}$ & \multicolumn{1}{r}{26.09 M} & \multicolumn{1}{r}{19.22 G} & \multicolumn{1}{r|}{183} & 0.6100$_{\pm 0.0017}$ & 0.9646$_{\pm 0.0003}$ \\
\midrule

With scaling & \multicolumn{1}{r}{17.17 M} & \multicolumn{1}{r}{17.49 G} & \multicolumn{1}{r|}{245} & 0.6170$_{\pm 0.0001}$ & 0.9664$_{\pm 0.0003}$ \\
Uni-Conv/Uni-SSM & \multicolumn{1}{r}{15.90 M} & \multicolumn{1}{r}{16.03 G} & \multicolumn{1}{r|}{414} & 0.6280$_{\pm 0.0010}$ & 0.9680$_{\pm 0.0008}$ \\
Uni-Conv/Bi-SSM & \multicolumn{1}{r}{16.10 M} & \multicolumn{1}{r}{16.43 G} & \multicolumn{1}{r|}{265} & 0.6277$_{\pm 0.0048}$ & 0.9687$_{\pm 0.0009}$ \\
Noam & \multicolumn{1}{r}{17.17 M} & \multicolumn{1}{r}{17.49 G} & \multicolumn{1}{r|}{254} & 0.6115$_{\pm 0.0045}$ & 0.9654$_{\pm 0.0008}$ \\

ECG-Mamba-V2$^{t}$ & \multicolumn{1}{r}{0.58 M} & \multicolumn{1}{r}{0.07 G} & \multicolumn{1}{r|}{6210} & 0.5420$_{\pm 0.0035}$ & 0.9453$_{\pm 0.0005}$ \\
Tiny-Mid-CLS & \multicolumn{1}{r}{0.58 M} & \multicolumn{1}{r}{0.07 G} & \multicolumn{1}{r|}{6210} & 0.5368$_{\pm 0.0006}$ & 0.9448$_{\pm 0.0003}$ \\
ECG-Mamba-V2$^{t,d}$ & \multicolumn{1}{r}{0.58 M} & \multicolumn{1}{r}{0.07 G} & \multicolumn{1}{r|}{6108} & 0.5448$_{\pm 0.0035}$ & 0.9464$_{\pm 0.0010}$ \\
ECG-Mamba-V2$^{t,ud}$ & \multicolumn{1}{r}{0.58 M} & \multicolumn{1}{r}{0.07 G} & \multicolumn{1}{r|}{6160} & 0.5376$_{\pm 0.0027}$ & 0.9444$_{\pm 0.0006}$ \\

ECG-Mamba-V2 & \multicolumn{1}{r}{17.17 M} & \multicolumn{1}{r}{17.49 G} & \multicolumn{1}{r|}{254} & 0.6333$_{\pm 0.0073}$ & 0.9691$_{\pm 0.0008}$ \\
Mid-CLS & \multicolumn{1}{r}{17.17 M} & \multicolumn{1}{r}{17.49 G} & \multicolumn{1}{r|}{254} & 0.6062$_{\pm 0.0060}$ & 0.9639$_{\pm 0.0018}$ \\
ECG-Mamba-V2$^{d}$ & \multicolumn{1}{r}{17.17 M} & \multicolumn{1}{r}{17.49 G} & \multicolumn{1}{r|}{254} & 0.6436$_{\pm 0.0009}$ & 0.9710$_{\pm 0.0007}$ \\
\textbf{ECG-Mamba-V2$^{ud}$} & \multicolumn{1}{r}{17.17 M} & \multicolumn{1}{r}{17.49 G} & \multicolumn{1}{r|}{252} & \textbf{0.6494}$_{\pm 0.0055}$ & \textbf{0.9716}$_{\pm 0.0007}$ \\

\bottomrule
\end{tabular}%
}
\end{table}

\begin{table}
\centering
\caption{Performance comparison on PhysioNet/CinC Challenge 2020. Abbreviations and superscripts as in Table~\ref{table_ECG-mamba-V2-2021}. Values are the mean over 15 runs (3 seeds $\times$ 5 folds) with the standard deviation in subscript.}
\label{table_ECG-mamba-V2}
\resizebox{\columnwidth}{!}{
\begin{tabular}{llll|ll}
\toprule
\textbf{Method} & \textbf{Params} & \textbf{MACs} & \textbf{T-put (f/s)} & \textbf{AUPRC} & \textbf{AUROC} \\ 
\midrule
Prna & \multicolumn{1}{r}{13.64 M} & \multicolumn{1}{r}{2.62 G} & \multicolumn{1}{r|}{2282} & 0.5005$_{\pm 0.0034}$ & 0.9279$_{\pm 0.0005}$ \\
ECG-Mamba$^{d}$ & \multicolumn{1}{r}{26.09 M} & \multicolumn{1}{r}{19.22 G} & \multicolumn{1}{r|}{184} & 0.5554$_{\pm 0.0038}$ & 0.9524$_{\pm 0.0017}$ \\
\textbf{ECG-Mamba-V2$^{ud}$} & \multicolumn{1}{r}{17.17 M} & \multicolumn{1}{r}{17.49 G} & \multicolumn{1}{r|}{254} & \textbf{0.5681}$_{\pm 0.0099}$ & \textbf{0.9561}$_{\pm 0.0022}$ \\
\bottomrule
\end{tabular}
}
\end{table}

The ablations isolate each design choice. Mid-CLS is the most damaging, reducing AUPRC from 0.6333 to 0.6062 at identical cost; substituting the Noam schedule costs a comparable margin (0.6115), and retaining the scaling factor yields 0.6170. Dropout contributes 0.0161, the uniform rate outperforming the linearly increasing one (0.6494 against 0.6436). Among three encoder configurations \cite{erol2024audio} --- Uni-Conv/Uni-SSM, Uni-Conv/Bi-SSM, and the full bidirectional encoder (ECG-Mamba-V2) --- SSM directionality contributes nothing (0.6280 against 0.6277) while convolution directionality accounts for the gain to 0.6333, though the margin lies within run-to-run variation and Uni-Conv/Uni-SSM runs 63\% faster (414 against 254 f/s).

The tiny configuration ECG-Mamba-V2$^{t,d}$ (4 blocks, model dimension 128, input length 1000, matching the 2D-CNN baseline so that both process identical input dimensions) equals that baseline in parameter count (0.58 M against 0.59 M) while requiring two orders of magnitude fewer MACs (0.07 G against 7.47 G), running 24\% faster (6108 against 4913 f/s), and exceeding ISIBrno in AUPRC (0.5448 against 0.5170) at the accuracy of the 2D-CNN (0.5431). The class token finding replicates at this scale: Tiny-Mid-CLS reaches 0.5368 against 0.5420 for ECG-Mamba-V2$^{t}$, identical apart from token position.

Table~\ref{table_ECG-mamba-V2} repeats the comparison on Challenge 2020, where ECG-Mamba-V2$^{ud}$ again leads at 0.5681 macro AUPRC and 0.9561 macro AUROC, exceeding ECG-Mamba$^{d}$ by 0.0127 and 0.0037 at the same cost advantage. Both state space models exceed Prna by a wide accuracy margin (0.5005), though Prna is far cheaper to run. The margin is narrower and the variation larger ($\pm$0.0099 against $\pm$0.0038) than on Challenge 2021; under the same test the model wins 12 of the 15 paired runs (median 0.0120 and 0.0034; $W = 10$, $p = 0.003$). As the 2021 dataset is a superset of the 2020 dataset and roughly twice its size, Challenge 2021 is the more reliable indicator, with Challenge 2020 serving as a secondary benchmark.

\section{Conclusion}
This letter presented ECG-Mamba-V2, a set of architectural refinements to a bidirectional Vision Mamba encoder for multi-label 12-lead ECG classification. The contribution is empirical: it identifies which design choices of the original ECG-Mamba actually drive the performance gains and which do not.

On the PhysioNet/CinC Challenge 2021 dataset, ECG-Mamba-V2\mbox{$^{ud}$} reached 0.6494 macro AUPRC and 0.9716 macro AUROC against 0.6100 and 0.9646 for ECG-Mamba\mbox{$^{d}$}, winning all 15 paired runs (\mbox{$p = 6.1 \times 10^{-5}$}), and exceeded the ISIBrno (0.5170, 0.9073) and 2D-CNN (0.5431, 0.9389) baselines. The 2020 dataset showed the same ordering, with 0.5681 and 0.9561 against 0.5005 and 0.9279 for Prna. These gains were obtained while reducing parameters by 34\% (17.17 M against 26.09 M) and raising throughput by 38\% over the predecessor, and the tiny configuration retains competitive accuracy at 0.07 G MACs. The ablations further show that class token placement and the learning rate schedule matter more than encoder bidirectionality at this scale. 
These findings indicate that SSM-based encoders are worth further investigation for 12-lead ECG classification.

\fontsize{8.5pt}{10pt}\selectfont




\section*{Competing Interests}
The authors declare no competing financial or non-financial interests. 
\bibliographystyle{fcs}
\bibliography{ref}


\end{document}